\pdfoutput=1

\documentclass{article}

\usepackage[preprint]{neurips_2025}

\usepackage[linesnumbered,ruled,vlined]{algorithm2e}
\usepackage{enumitem}
\usepackage{arydshln} 
\usepackage{amssymb}
\usepackage[utf8]{inputenc} 
\usepackage[T1]{fontenc}    
\usepackage{url}            
\usepackage{booktabs}       
\usepackage{amsfonts}       
\usepackage{nicefrac}       
\usepackage{microtype}      
\usepackage[table]{xcolor}          
\usepackage{graphicx}
\usepackage{booktabs}
\usepackage{multirow}
\usepackage{bbding}
\usepackage{amsmath}
\usepackage{orcidlink}

\usepackage{makecell}
\usepackage{subcaption}  
\usepackage{graphicx}
\usepackage{lipsum}    
\usepackage{wrapfig}
\usepackage[linesnumbered,ruled,vlined]{algorithm2e}

\title{SCA: Improve Semantic Consistent in Unrestricted Adversarial Attacks via DDPM Inversion}

\author{
    Zihao Pan\textsuperscript{1}, 
    Lifeng Chen\textsuperscript{2}, 
    Weibin Wu\textsuperscript{1}\thanks{Corresponding Author}, 
    Yuhang Cao\textsuperscript{1}, 
    Zibin Zheng\textsuperscript{1} \\
    \textsuperscript{1}School of Software Engineering, Sun Yat-sen University \\
    \textsuperscript{2}Computer Science, Beijing Jiaotong University \\
}

\begin{document}

\maketitle

\begin{abstract}
Systems based on deep neural networks are vulnerable to adversarial attacks. Unrestricted adversarial attacks typically manipulate the semantic content of an image (e.g., color or texture) to create adversarial examples that are both effective and photorealistic. Recent works have utilized the diffusion inversion process to map images into a latent space, where high-level semantics are manipulated by introducing perturbations. However, they often result in substantial semantic distortions in the denoised output and suffer from low efficiency.  In this study, we propose a novel framework called Semantic-Consistent Unrestricted Adversarial Attacks (SCA), which employs an inversion method to extract edit-friendly noise maps and utilizes a Multimodal Large Language Model (MLLM) to provide semantic guidance throughout the process. Under the condition of rich semantic information provided by MLLM, we perform the DDPM denoising process of each step using a series of edit-friendly noise maps and leverage DPM Solver++ to accelerate this process, enabling efficient sampling with semantic consistency. Compared to existing methods, our framework enables the efficient generation of adversarial examples that exhibit minimal discernible semantic changes. Consequently, we for the first time introduce Semantic-Consistent Adversarial Examples (SCAE). Extensive experiments and visualizations have demonstrated the high efficiency of SCA, particularly in being on average 12 times faster than the state-of-the-art attacks. Our code can be found at \url{https://github.com/Pan-Zihao/SCA}.
  
\end{abstract}

\section{Introduction}
Deep neural networks have excelled in learning features and achieving impressive performance across various tasks. However, they are vulnerable to small perturbations, known as adversarial samples, which raise significant security concerns for critical decision-making systems \cite{9727149,10714478,9663154,10388391,8482346,10602524,10035539,8854834}. With the explosive growth of digital information, they have triggered security risks such as face recognition \cite{9116802,9252132}, voice recognition \cite{8747397,9205635,9954062} and information forensics \cite{9495178,9773023}. 

\begin{figure}[t]
\centering
\subcaptionbox{Reconstruction Result and Adversarial Example of Our Method}{
\subcaptionbox*{Clean Image}{\includegraphics[width = 0.3\textwidth]{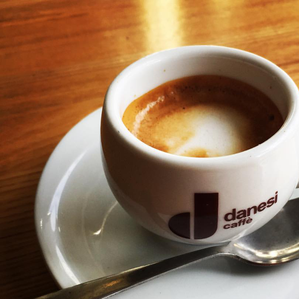}}
\hfill
\subcaptionbox*{Reconstruction Result}{\includegraphics[width = 0.3\textwidth]{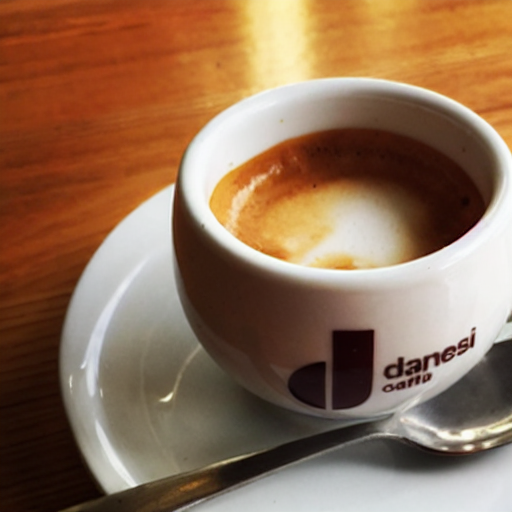}}
\hfill
\subcaptionbox*{Adversarial Example}{\includegraphics[width = 0.3\textwidth]{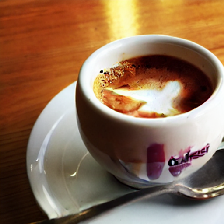}}
}
\\
\subcaptionbox{Comparison of Our Method with the SOTA Approach}{
\subcaptionbox*{Clean Image}{\includegraphics[width = 0.3\textwidth]{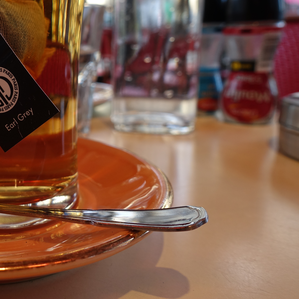}}
\hfill
\subcaptionbox*{Previous SOTA}{\includegraphics[width = 0.3\textwidth]{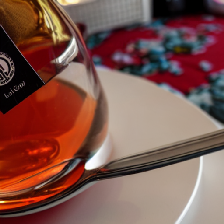}}
\hfill
\subcaptionbox*{\textbf{Ours}}{\includegraphics[width = 0.3\textwidth]{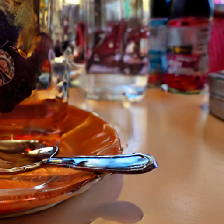}}
}
\caption{The above shows the superiority of SCA in maintaining semantic consistency. (a) shows that our method achieves perfect reconstruction and introduces minimal perturbations later. It can be seen that we only made some changes to the text on the coffee cup and the coffee pattern, and the overall semantics and environment remained consistent. This shows that our method can accurately identify the subject in the image and attack it without affecting the overall image, greatly improving its concealment from human perception. (b) is a comparison between the current state-of-the-art method called ACA and SCA. Compared with the adversarial examples generated by SCA, ACA causes a large semantic deviation in the image.}
\label{fig:1}
\end{figure}

\begin{figure}[t]
\begin{center}
\subcaptionbox{Loss of Original Image Information}{
    \includegraphics[width = 0.23\textwidth]{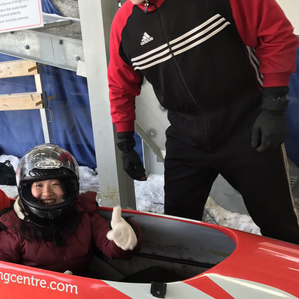}
    \includegraphics[width = 0.23\textwidth]{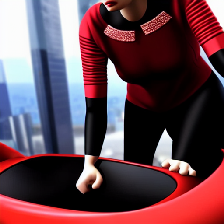}
    }
\hfill
\subcaptionbox{Artifacts in Generated Images}{
    \includegraphics[width = 0.23\textwidth]{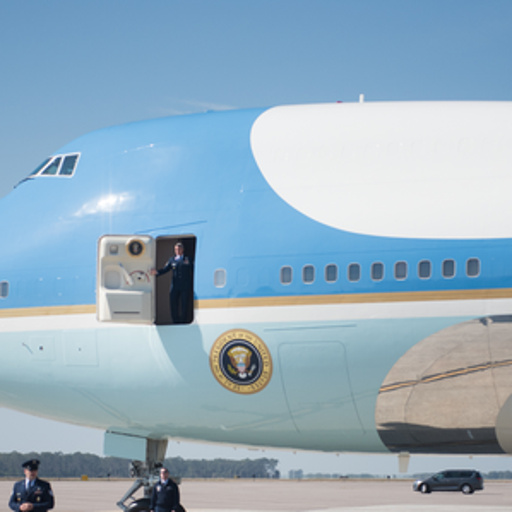} 
    \includegraphics[width = 0.23\textwidth]{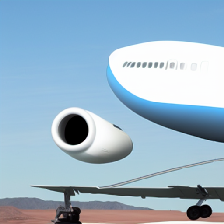}
    }
\end{center}
\caption{The above shows some examples of failure of the existing state-of-the-art methods. We can see that although it can generate rich and diverse content with the help of text-to-image models, large semantic deviations often exist. The image on the left is clean, and the image on the right is a generated adversarial example. We can see that the person's information is lost in (a). The original image is described as ``a woman sitting in a sled and a man standing beside her''. However, the generated result loses the information of the person sitting in the sled. In addition, the object structure is abnormal in (b).}
\label{fig:add}
\end{figure}

To ensure human visual imperceptibility and maintain photorealism, adversarial examples must retain high semantic consistency with the original images, appearing natural and realistic despite minor perturbations \cite{hosseini2018semantic,liu2023towards,xiang2021improving,zhao2018generating}. A recent trend in adversarial attacks is the emergence of unrestricted adversarial attacks \cite{bhattad2019unrestricted,brown2018unrestricted,liu2023towards,song2018constructing}, which alter semantic features such as color and texture to produce adversarial examples that are both visually plausible and highly effective \cite{hosseini2018semantic,liu2024generation}. State-of-the-art methods \cite{chen2024content,xu2024highly} utilize generative models like Stable Diffusion \cite{rombach2022high}, leveraging their knowledge of natural image distributions to map images onto a latent space, where small perturbations are introduced to optimize the adversarial objective. While this approach allows for more flexible adversarial examples, it often causes noticeable semantic shifts, reducing stealthiness to human observers \cite{lau2023attribute,zhao2017generating}. A comparison of our method with previous work on semantic consistency is shown in Figure~\ref{fig:1}. We argue that even small changes in the latent space can result in significant image alterations. Existing methods, constrained by \(l_p\) norms, struggle to control these semantic variations, as shown in Figure~\ref{fig:add}. Furthermore, they typically require extensive time steps (e.g., 1000) for accurate image inversion, making the attack process both time-consuming and prone to uncontrolled semantic changes \cite{chen2024content,xu2024highly}.

To tackle the challenges inherent in adversarial attacks, we introduce a novel approach called Semantic-Consistent Unrestricted Adversarial Attack (SCA), which leverages rich semantic information from the original images as a guide throughout the entire attack process. Our method begins by enhancing the inversion mechanism. Denoising Diffusion Probabilistic Models (DDPMs) typically produce images by progressively refining a series of white Gaussian noise samples. Departing from prior approaches \cite{chen2023advdiffuser,dai2023advdiff,xiang2021improving} we propose an alternative latent noise space for DDPM, which allows for a wide range of editable operations through straightforward manipulation. This space enables the extraction of noise maps that are conducive to editing for a given image. In the context of generating unrestricted adversarial examples, an essential property of inversion is that perturbing the latent space while keeping the noise maps fixed should result in artifact-free images that retain a high degree of semantic consistency with the original image. To achieve this, the noise maps are designed to be statistically independent and to follow a standard normal distribution, akin to conventional sampling techniques. Our inversion process imprints the image more strongly onto the noise maps, thereby preserving semantics more effectively when traversing the low-dimensional manifold. This preservation is because edit-friendly noise maps exhibit higher variances than their native counterparts. Additionally, Multimodal Large Language Models (MLLMs) have demonstrated a strong capacity for understanding a wide array of image semantics \cite{yin2023survey}. We leverage MLLMs to generate highly detailed image descriptions, which guide both the inversion and generation processes, thereby providing rich prior knowledge to maintain semantic consistency during perturbation in the latent space. Finally, we integrate DPM Solver++ \cite{lu2022dpm++}, an optimized numerical algorithm that accelerates the sampling process of DDPMs by reducing the number of required time steps to just 10–20. This significantly boosts the overall efficiency of our method.

In summary, our main contributions are as follows:

\begin{itemize} \item We introduce a novel attack framework, \textbf{Semantic-Consistent Unrestricted Adversarial Attack}, which leverages an effective inversion method and powerful MLLM to generate adversarial examples with minimal semantic distortion. \item We develop a more covert and natural unrestricted adversarial attack through \textbf{Semantic Fixation Inversion} and \textbf{Semantically Guided Perturbation}. This results in \textbf{Semantic-Consistent Adversarial Examples (SCAE)}, which maintain content diversity while deceiving DNN models without altering the image's overall semantics, making them harder to detect and defend against. Our approach provides insights into DNN vulnerabilities and informs novel defense strategies. \item We demonstrate the effectiveness of our attack through comprehensive experiments and visual analyses. We achieve comparable attack success rates and transferability to current state-of-the-art methods, with a 12x improvement in efficiency and enhanced semantic consistency. \end{itemize}

\section{Related Work}
Adversarial attacks aim at maximizing the classification error of a target model without changing the semantics of the images. In this section, we review the types of attacks and existing works.

\textbf{Norm-Bounded Attacks.} Norm-bounded attacks are popular for their simplicity and effectiveness, using gradient-based optimization to maximize model loss \cite{9727149,10035539,10602524}. Techniques like FGSM \cite{goodfellow2014explaining}, L-BFGS \cite{szegedy2013intriguing}, PGD \cite{madry2017towards}, and the CW attack \cite{carlini2017towards} constrain perturbations within an $l_p$ norm, ensuring minimal modifications. However, these methods often introduce distortions detectable by humans and robust models, making them less effective in real-world applications. Norm constraints also limit flexibility, as $l_p$ norms do not always align with human perception, leading to visually suboptimal results.

\begin{figure*}[t]
\begin{center}
   \includegraphics[width=\linewidth]{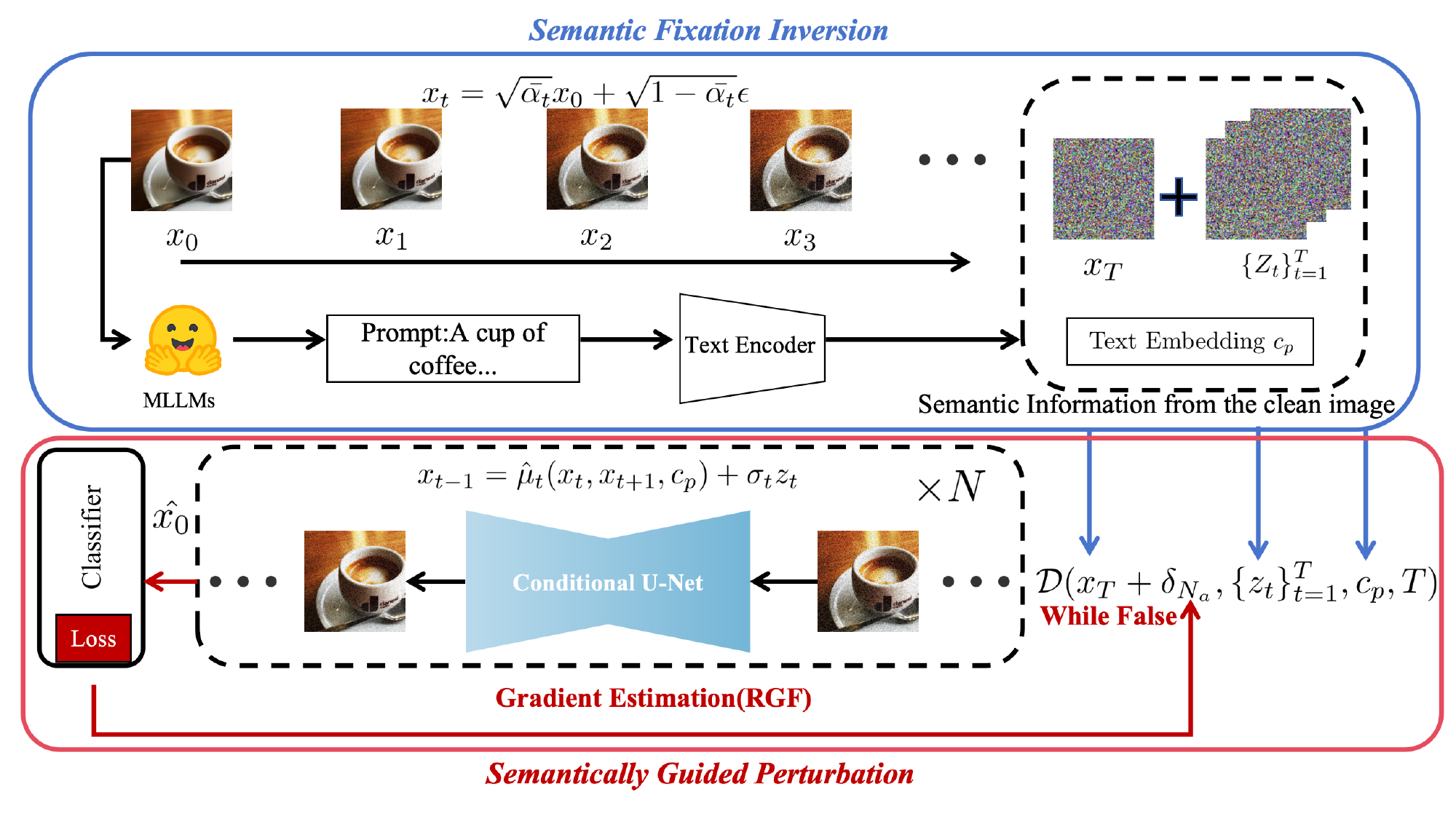}
\end{center}
   \caption{Pipeline of Semantic-Consistent Unrestricted Adversarial Attack. We first map the clean image into a latent space through Semantic Fixation Inversion, and then iteratively optimize the adversarial objective in the latent space under semantic guidance, causing the content of the image to shift in the direction of deceiving the model until the attack is successful.}
\label{fig:2}
\end{figure*}

\textbf{Unrestricted Adversarial Attacks.} 
As limitations of normbounded attacks become increasingly apparent, a growing
interest in unrestricted adversarial attacks has emerged. These
attacks aim to replace the traditional small $l_p$ norm perturbations with unrestricted but natural transformations, making
them more applicable in real-world scenarios. Unrestricted
adversarial attacks typically rely on semantic transformations,
such as changes in shape, texture, or color, to craft adversarial
examples. Shape-based unrestricted attacks, such as those by
Alaifari et al. \cite{alaifari2018adef} and Xiao et al. \cite{xiao2018spatially}, iteratively introduce
minor deformations to an image using gradient descent to
generate adversarial examples. These methods exploit subtle
changes in the geometry of the image but are often limited
by the complexity of maintaining natural-looking outcomes.
Following shape-based approaches, texture-based attacks
emerged as an alternative. Bhattad et al. \cite{bhattad2019unrestricted}, Qiu et al. \cite{qiu2020semanticadv} and Mou et al. \cite{10714478} proposed methods that manipulate an image’s texture or style
to induce adversarial behavior. While texture-based attacks
provide an effective means of altering perceptual features, they
often lead to visually unnatural results, thereby diminishing their practical utility and adversarial transferability
across different models. The unnatural distortions generated by these approaches highlight the challenge of balancing
perceptual realism with adversarial effectiveness.

To address these challenges, researchers turned to color-based unrestricted attacks, which aim to manipulate pixel
values along specific dimensions to maintain a higher degree of
realism. Works by Hosseini et al. \cite{hosseini2018semantic}, Bhattad et al. \cite{bhattad2019unrestricted}, Zhao
et al. \cite{zhao2020adversarial}, Shamsabadi et al. \cite{shamsabadi2020colorfool}, and Yuan et al. \cite{yuan2022natural} focus on
altering color properties to generate adversarial examples that
appear more natural to human observers. Despite achieving
photorealism, these attacks often sacrifice flexibility in generating diverse adversarial examples. The reliance on subjective
visual intuitions, coupled with the use of objective metrics
like perceptual similarity, constrains the range of possible
adversarial transformations. Consequently, color-based attacks
frequently exhibit limited transferability to other models, reducing their effectiveness in broader adversarial settings.

In summary, while shape-, texture-, and color-based unrestricted attacks have contributed significantly to the field
of adversarial machine learning, each method comes with
inherent trade-offs between realism, flexibility, and adversarial
transferability. Our work builds on these insights, aiming to
overcome the limitations of existing approaches by developing
a more flexible and transferable framework for generating natural adversarial examples without compromising photorealism.
 
\textbf{Latent Space-Based Attacks.} 
Recent advancements in
adversarial attack techniques have focused on manipulating the
latent space of generative models to craft adversarial images.
For instance, Zhao et al.\cite{zhao2017generating}, Lin et al. \cite{lin2020dual}, Hu et al. \cite{hu2021naturalistic},
Lapid et al. \cite{lapid2023patch}, and Lau et al. \cite{lau2023attribute} leveraged Generative
Adversarial Networks (GANs), fine-tuning the generator to
produce adversarial images. These methods typically adjust
the generator’s parameters to subtly modify the image’s latent
representation, thus creating visually plausible yet adversarial
content. On the other hand, diffusion models have also been
employed for similar purposes. Xue et al. \cite{xue2024diffusion}, Wang et
al. \cite{wang2023semantic}, Chen et al. \cite{chen2023advdiffuser}, Liu et al. \cite{liu2023diffprotect}, and Chen et al. \cite{chen2024content} focused on optimizing the U-Net structure or injecting
learned noise into the latent space to generate adversarial examples. Although diffusion models offer enhanced flexibility
in controlling image generation processes, a common issue
across both GAN-based and diffusion-based approaches is the
pronounced semantic shift in the adversarial examples. These
shifts can often make the manipulations more perceptible to
humans, thus undermining the stealthiness of the attack. In
contrast, our proposed framework addresses these limitations
by generating adversarial examples that maintain a higher
degree of semantic consistency, ensuring the resulting images
exhibit minimal perceptible alterations while still achieving
the desired adversarial effect. This improvement in stealthiness
enhances the practical viability of adversarial attacks in real-world applications.

\IncMargin{1em}
\begin{algorithm}[t] 
\SetKwFunction{Union}{Union}\SetKwFunction{FindCompress}{FindCompress} \SetKwInOut{Input}{input}\SetKwInOut{Output}{output}
	
	\Input{A input image $x_0$ with the label $y$, a text embedding $c_p=\psi(\mathcal{P})$, a classifier $\mathcal{F}_{\theta}(\cdot)$, DDPM steps $T$, attack iterations $N_a$, momentum factor $\mu$, and $N$ is the number of queries.} 
	\Output{Semantic-Consistent Unrestricted Adversarial Example $\hat{x_0}$}
	 \BlankLine 
	 
	 \emph{//\textbf{\textit{Semantic Fixation Inversion}}}\; 
	 \For{$t=1 \to T$}{ 
	 	$\tilde{\epsilon}\sim\mathcal{N}(0, 1)$\; 
	 	$x_{t}\leftarrow\sqrt{\bar{\alpha}_{t}}x_{0}+\sqrt{1-\bar{\alpha}_{t}}\tilde{\epsilon}$\; }
        \For{$t=T-1 \to 1$}{
        $z_t\leftarrow\frac{x_{t-1}-\hat{\mu}_t(x_t,x_{t+1},c_p)}{\sigma_t}$\;
        $x_{t-1}\leftarrow\hat{\mu_t}(x_t,x_{t+1},c_p)+\sigma_t z_t$\;
        }
       \emph{//\textbf{\textit{Semantically Guided Perturbation}}}\;
	 	\For{$k=1,\ldots,N_a$}{
        $\hat{x_0} \gets \mathcal{D}(x_T,\{z_t\}_{t=1}^{T},c_p,T)$\;
        $g_{k}\leftarrow\mu\cdot g_{k-1}+\frac{\nabla_{z_{T}}\mathcal{L}\left(\mathcal{F}_{\theta}\left((\varrho(\hat{x_{0}}),y)\right)\right)}{||\nabla_{z_{T}}\mathcal{L}\left(\mathcal{F}_{\theta}\left(\varrho(\hat{x_{0}}),y\right)\right)||_{1}}$\;
        $\delta_k\leftarrow\Pi_\kappa\left(\delta_{k-1}+\eta\:\cdot\mathrm{sign}(g_k)\right)$\;
        }
        $\hat{x_0} \gets \varrho(\mathcal{D}(x_T+\delta_{N_a},\{z_t\}_{t=1}^{T},c_p,T))$\;
 	 	  \caption{\textbf{SCA}}
 	 	  \label{al1}
 	 \end{algorithm}
 \DecMargin{1em} 

\section{Preliminaries}
Diffusion models (DMs) generate data by iteratively denoising a Gaussian-distributed random variable to approximate a learned data distribution through forward and reverse processes. In the forward process, an image $x_0$ is transformed into Gaussian noise via:

\begin{equation}
    x_{t} = \sqrt{1 - \beta_t} \, x_{t-1} + \sqrt{\beta_t} \, n_t, \quad t = 1, \dots, T,
    \label{eq:2}
\end{equation}
where $n_t \sim \mathcal{N}(0, \mathbf{I})$ is noise, and $\beta_t$ controls the noise injection. This leads to $x_T \sim \mathcal{N}(0, \mathbf{I})$. Marginalizing over intermediate states gives:

\begin{equation}
    x_t = \sqrt{\bar{\alpha}_t} \, x_0 + \sqrt{1 - \bar{\alpha}_t} \, \epsilon_t,
    \label{eq:3}
\end{equation}
with $\epsilon_t \sim \mathcal{N}(0, \mathbf{I})$. In the reverse process, the model denoises $x_T$ to reconstruct $\hat{x}_0$ as:

\begin{equation}
    x_{t-1} = \hat{\mu}_t(x_t) + \sigma_t z_t, \quad t = T, \dots, 1,
    \label{eq:4}
\end{equation}
where $\sigma_t = \eta \beta_t \frac{1 - \bar{\alpha}_{t-1}}{1 - \bar{\alpha}_t}$ controls stochasticity. The mean prediction is:

\begin{equation}
\begin{aligned}
    \hat{\mu}_t(x_t) = &\sqrt{\bar{\alpha}_{t-1}} \frac{x_t - \sqrt{1 - \bar{\alpha}_t} \, \hat{\epsilon}_\theta(x_t)}{\sqrt{\bar{\alpha}_t}} \\
    &+ \sqrt{1 - \bar{\alpha}_{t-1} - \sigma_t^2} \, \hat{\epsilon}_\theta(x_t).
\end{aligned}
\end{equation}

For a text-conditioned generation, classifier-free guidance adjusts the noise prediction:

\begin{equation}
    \hat{\epsilon}_\theta(x_t, c_p) := \hat{\epsilon}_\theta(x_t) + s_g \left( \hat{\epsilon}_\theta(x_t, c_p) - \hat{\epsilon}_\theta(x_t) \right),
    \label{eq:5}
\end{equation}
with $s_g$ balancing conditional and unconditional objectives, enabling high-quality, text-guided image generation.

\section{method}

The pipeline of our proposed method is shown in Figure~\ref{fig:2}. The core idea of Semantic-Consistent Unrestricted Adversarial Attack is to enhance semantic control throughout the entire generation process of unrestricted adversarial examples, which is achieved by introducing a novel inversion method that can ``imprint'' the image more strongly onto the noise maps. Furthermore, the powerful semantic guidance provided by MLLMs further restricts the direction of perturbations in the latent space, enabling the clean image to produce our desired adversarial changes, namely imperceptibility, and naturalness.

Specifically, our method can be divided into two parts: \textbf{Semantic Fixation Inversion} and \textbf{Semantically Guided Perturbation}. We first map the clean image into a latent space through Semantic Fixation Inversion, and then iteratively optimize the adversarial objective in the latent space under semantic guidance, causing the content of the image to shift in the direction of deceiving the model until the attack is successful. The algorithm of SCA is presented in Algorithm~\ref{al1}.

\subsection{Problem Definition.} For a deep learning classifier $\mathcal{F}\theta(\cdot)$ with parameters $\theta$, let $x$ represent the clean image and $y$ the corresponding ground truth label. Unrestricted adversarial attacks seek to introduce subtle adversarial perturbations (e.g., image distortions, texture, or color alterations, etc.) to the input $x$, resulting in an adversarial example $x{adv}$ that misguides the classifier $\mathcal{F}_\theta(\cdot)$:
\begin{equation}
    \max_{x_{adv}}\mathcal{L}(\mathcal{F}_{\theta}(x_{adv}),y),\quad s.t.\:d(x,x_{adv})<\epsilon\label{eq:1},
\end{equation}
where $\mathcal{L}(\cdot)$ is the loss function and $d(x,x_{adv})$ represents the semantic distance between $x$ and $x_{adv}$. Based on the above, we believe that ideal unrestricted adversarial examples should have high semantic consistency.

\subsection{Semantic Fixation Inversion}
In the context of Semantic Fixation Inversion, the key objective is to extract rich and semantically relevant information from the original image during the inversion process. This semantic information acts as an informative prior, guiding both the perturbation of the latent space and the image reconstruction, facilitating a Semantic Consistency Attack. In our method, the semantic priors are derived from edit-friendly noise maps and the powerful capabilities of Multimodal Language Models (MLLMs). Recent research predominantly focuses on inverting the diffusion sampling process to determine $x_T$, which can be subsequently denoised into the input image $x_0$. Among the inversion techniques, inverting the DDPM (Denoising Diffusion Probabilistic Models) scheduler is often favored over DDIM (Denoising Diffusion Implicit Models) inversion, as the former can achieve results in fewer steps while maintaining perfect reconstruction \cite{huberman2024edit}.

Although the DDPM scheduler provides efficient inversion, more advanced approaches exist for sampling in Diffusion Models (DMs), significantly reducing the number of steps and, consequently, the number of DM evaluations. In our work, we adopt a more efficient inversion approach \cite{brack2024ledits++}, which retains the essential inversion properties while accelerating the process. Previous work by Song et al. \cite{songscore} has shown that the reverse diffusion process in DDPM can be viewed as a first-order stochastic differential equation (SDE). Solving this SDE with higher-order methods, such as the dpm-solver++ \cite{lu2022dpm++}, enables more efficient computation and reduces the number of required steps.

Leveraging the semantic understanding capability of MLLMs, which have been shown to provide accurate semantic interpretations across diverse image domains, we employ MLLMs to generate rich captions \( \mathcal{P} \) for the clean input image \( x_0 \). These captions are then converted into conditional embeddings \( c_p \) via a function \( \psi(\mathcal{P}) \), serving as the conditional input for the pre-trained diffusion model parameterized by \( \hat{\epsilon}_{\theta} \). This process integrates seamlessly with the existing formulation for conditional noise estimation \( \hat{\epsilon}_{\theta}(x_t, c_p) \) as given in Equation~(\ref{eq:5}). 

The reverse diffusion process, which uses the second-order dpm-solver++ SDE solver, can be written as:

\begin{equation}
    x_{t-1} = \hat{\mu}_t(x_t, x_{t+1}, c_p) + \sigma_t z_t, \quad t = T, \dots, 1.
    \label{eq:6}
\end{equation}

In this expression, the variance term $\sigma_t$ is now defined as:

\begin{equation}
    \sigma_t = \sqrt{1 - \bar{\alpha}_{t-1}} \, \sqrt{1 - e^{-2h_{t-1}}}.
\end{equation}

This updated variance scheduling leads to improved noise control, and the mean prediction \( \hat{\mu}_t(x_t, x_{t+1}, c_p) \) depends on information from both the current timestep $x_t$ and the subsequent one $x_{t+1}$, making the process higher-order. Specifically, \( \hat{\mu}_t(x_t, x_{t+1}, c_p) \) is given by:

\begin{equation}
\begin{aligned}
    &\frac{\sqrt{1 - \bar{\alpha}_{t-1}}}{\sqrt{1 - \bar{\alpha}_t}} e^{-h_{t-1}} x_t + \sqrt{\bar{\alpha}_{t-1}} \left( 1 - e^{-2h_{t-1}} \right) \hat{\epsilon}_{\theta}(x_t, c_p) \\
    + &0.5 \, \sqrt{\bar{\alpha}_{t-1}} \left( 1 - e^{-2h_{t-1}} \right) \frac{-h_t}{h_{t-1}} \left( \hat{\epsilon}_{\theta}(x_{t+1}, c_p) - \hat{\epsilon}_{\theta}(x_t, c_p) \right).
    \label{eq:7}
\end{aligned}
\end{equation}

The term \( h_t \) is a function of the variance schedule and is given by:

\begin{equation}
    h_t = \frac{\ln(\sqrt{\bar{\alpha}_t})}{\ln(\sqrt{1 - \bar{\alpha}_t})} - \frac{\ln(\sqrt{\bar{\alpha}_{t+1}})}{\ln(\sqrt{1 - \bar{\alpha}_{t+1}})}.
\end{equation}

This higher-order solver significantly improves efficiency by using information from two adjacent timesteps, allowing for faster convergence without sacrificing reconstruction quality. For more comprehensive information on the solver's derivation and its convergence behavior, we direct readers to the pertinent literature \cite{lu2022dpm++}. 

Based on the inversion process, we generate an auxiliary sequence of noise images beginning with the input image $x_0$, which can be formulated as follows:

\begin{equation}
    x_t = \sqrt{\bar{\alpha}_t} x_0 + \sqrt{1 - \bar{\alpha}_t} \tilde{\epsilon}_t,
    \label{eq:8}
\end{equation}
where \( \tilde{\epsilon}_t \sim \mathcal{N}(0, \mathbf{I}) \) are statistically independent noise vectors. In contrast to the correlated noise in Equation~(\ref{eq:3}), the independence of \( \tilde{\epsilon}_t \) is a desirable property for ensuring semantic consistency during the reconstruction process \cite{huberman2024edit}.

Finally, the noise term \( z_t \) required for the inversion process is derived from Equation~(\ref{eq:6}) as:

\begin{equation}
    z_t = \frac{x_{t-1} - \hat{\mu}_t(x_t, x_{t+1}, c_p)}{\sigma_t}, \quad t = T, \dots, 1,
    \label{eq:9}
\end{equation}
where \( \hat{\mu}_t \) and \( \sigma_t \) are as defined earlier. 

In conclusion, we have successfully derived a series of edit-friendly noise maps \( \{ x_T, z_1, z_2, \dots, z_T \} \), which can be progressively denoised to reconstruct the original image using the reverse diffusion process outlined in Equation~(\ref{eq:6}). The statistical independence of these noise maps, combined with the rich semantic priors provided by MLLMs, ensures that the reconstruction process maintains semantic consistency without unnecessary perturbations to the latent space. The subsequent experimental section provides a detailed analysis of the reconstruction quality and compares our results with existing methods. Additionally, the extracted semantic knowledge is employed to guide Semantically Guided Perturbation, enabling further manipulation of the generated images.

\subsection{Semantically Guided Perturbation}
In this section, we introduce an optimization technique for latents aimed at enhancing the attack performance on unrestricted adversarial examples. Drawing from previous work \cite{chen2024content}, perturbing the text embedding \( c_p \) to generate adversarial content is suboptimal because it is crucial for semantic consistency, preserving a wealth of semantic information. Additionally, due to the discrete nature of textual representations, current techniques struggle to generate smooth and diverse perturbations based on gradient methods. Consistent with state-of-the-art approaches\cite{chen2024content,xue2024diffusion,xu2024highly}, we opt to optimize the adversarial objective within the latent space. Building upon Semantic Fixation Inversion, we utilize rich semantic priors to guide the optimization process. We define the denoising process of diffusion models as \( \mathcal{D}(\cdot) \) through Equation~(\ref{eq:10}), involving \( T \) iterations:

\begin{equation}
    \begin{aligned}
\hat{x_0}  &= \hat{\mu}_1(\hat{x}_1, \hat{x}_2, c_p) + \sigma_1 z_1 \\
&= \hat{\mu}_1(\hat{\mu}_2(\hat{x}_2, \hat{x}_3, c_p) + \sigma_2 z_2, \hat{\mu}_3(\hat{x}_3, \hat{x}_4, c_p) + \sigma_3 z_3, c_p) \\&\quad+ \sigma_1 z_1 \\
&= \cdots \cdots \\ &= \mathcal{D}({x_T, z_1, z_2, \ldots, z_T}, c_p),
\label{eq:10}
\end{aligned}
\end{equation}
with 
\begin{equation}\hat{x_{t-1}}=\hat{\mu}_{t}(\hat{x_{t}},\hat{x_{t+1}},c_p)+\sigma_{t}z_{t},\quad t=T,...,1.
\end{equation}

Therefore, the reconstructed image is denoted as $\hat{z_0}=\mathcal{D}(x_T,\{z_t\}_{t=1}^{T},c_p,T)$.  The computational procedure of the VAE is omitted in this context, as it remains differentiable. Our adversarial objective is expressed by: 
\begin{equation}
\begin{split}
    \max _{\delta} \mathcal{L}&(\mathcal{F}_{\theta}(\hat{x_0}, y), \\ \text { s.t. }\|\delta\|_{\infty} \leq \kappa, &\hat{x_0}=\mathcal{D}(x_T,\{z_t\}_{t=1}^{T},c_p,T)\\ \text { and } d&(x,x_{adv})<\epsilon\ ,
    \label{eq:11}
\end{split}
\end{equation}
where $\delta$ is the adversarial perturbation on the latent space and $d$ represents the semantic distance. Previous works \cite{chen2024content,xu2024highly} typically design the loss function \( \mathcal{L} \) as comprising two components: i) the cross-entropy loss \( \mathcal{L}_{ce} \), which primarily guides adversarial examples towards misclassification, and ii) the mean square error loss \( \mathcal{L}_{mse} \), which mainly ensures that the generated adversarial examples are as close as possible to the clean image in terms of \( l_2 \) distance. However, extensive experiments have confirmed that the additional introduction of \( \mathcal{L}_{mse} \) in the loss function has a very limited role in semantic preservation \cite{chen2024content}, perhaps because pixel-level numerical changes are not a high-level semantic constraint. Furthermore, the introduction of two optimization objectives increases the difficulty of optimization and limits the capability of the attack. Specific experimental results regarding this component are presented in our ablation study. Since good semantic consistency has already been achieved during the Semantic Fixation Inversion phase, we let \( \mathcal{L} = \mathcal{L}_{ce} \).

Existing methods \cite{chen2024content,xu2024highly,xue2024diffusion} generally assume that a perturbation \( \delta \) will not disrupt semantic consistency when it is small, hence imposing a norm constraint on \( \delta \). However, we find that even small perturbations in the latent space can cause significant changes in the final image. Because the inversion process in these methods is approximate, it is not easy to precisely find the latent representation corresponding to the clean image. According to Equation~(\ref{eq:10}), our method avoids the accumulation of approximation errors in the optimization process by guiding the image-denoising process with the rich semantic knowledge excavated during the inversion phase. Analogous to conventional adversarial attacks \cite{dong2018boosting}, we employ gradient-based techniques to estimate \( \delta \) through \( \delta \simeq \eta \nabla_{x_{T}} \mathcal{L}\left(\mathcal{F}_{\theta}\left(\hat{x_{0}}\right), y\right) \), where \( \eta \) denotes the magnitude of perturbations that occur in the direction of the gradient. To expand \(\nabla_{x_{T}} \mathcal{L}\left(\mathcal{F}_{\theta}\left(\hat{x_{0}}\right), y\right) \) by the chain rule, we can obtain:
\begin{equation}
 \nabla_{x_{T}}\mathcal{L}\left(\mathcal{F}_{\theta}\left(\hat{x_{0}}\right), y\right)=\frac{\partial \mathcal{L} }{\partial \hat{x_0}} \frac{\partial \hat{x_0}}{\partial x_T} = \frac{\partial \mathcal{L} }{\partial \mathcal{D} }\frac{\partial \mathcal{D}}{\partial x_T}.
 \label{eq:12}
\end{equation}

\textbf{Gradient Estimation.} However, we cannot directly compute the gradients for optimization based on Equation~(\ref{eq:10}) and Equation~(\ref{eq:11}), as this is excessively complex and leads to memory overflow (similar phenomena are also observed in \cite{chen2024content,salman2023raising}). Previous works \cite{chen2024content,xu2024highly} have utilized the Skip Gradient technique to estimate this gradient, specifically by rearranging Equation~(\ref{eq:3}) to derive \( x_0 = \frac{1}{\sqrt{\bar{\alpha_t}}} x_T - \sqrt{1 - \bar{\alpha_t}} \epsilon \). Consequently, they further obtain \( \frac{\partial x_0}{\partial x_T} = \frac{1}{\sqrt{\bar{\alpha_T}}} \). However, this estimation method is overly simplistic and introduces significant errors. Additionally, our method differs from DDIM sampling as it is not an approximate method, especially given that we use fewer timesteps, which further magnifies the error.

To estimate the gradients, we employ the random gradient-free (RGF) method \cite{nesterov2017random}. First, we rewrite a gradient as the expectation of direction derivatives, i.e., $\nabla_{\boldsymbol{x}}F(\boldsymbol{x})=\mathbb{E}\left[\boldsymbol{\delta}^{\top}\nabla_{\boldsymbol{x}}F(\boldsymbol{x})\cdot\boldsymbol{\delta}\right]$, where $F(\boldsymbol{x})$ represents any differentiable function and $\delta\sim P(\boldsymbol{\delta})$ is a random variable satisfying that $\mathbb{E}[\boldsymbol{\delta\delta}^\top]=\mathbf{I}(e.g., \boldsymbol{\delta}$ can be uniformly sampled from a hypersphere). Then by zero-order optimization \cite{chen2017zoo,zhao2024evaluating}, we know that
\begin{equation}
\begin{split}
\nabla_{x_{T}}\mathcal{L}\left(\mathcal{F}_{\theta}\left(\hat{x_{0}}\right), y\right)&=\frac{\partial \mathcal{L} }{\partial \mathcal{D} }\frac{\partial \mathcal{D}}{\partial x_T}\\
\approx\frac{\partial \mathcal{L} }{\partial \mathcal{D} }\frac1{N\sigma}&\sum_{n=1}^N[\mathcal{D}(x_T+\sigma\boldsymbol{\delta}_n)-\mathcal{D}(x_T)]\cdot\boldsymbol{\delta}_n,
    \label{eq:13}
\end{split}
\end{equation}
where $\delta_n\sim P(\delta),\sigma$ is a hyperparameter controls the sampling variance, and $N$ is the number of queries. The approximation in  Equation~(\ref{eq:13}) becomes an unbiased equation when $\sigma\to0$ and $N\to\infty.$

\textbf{Differentiable Boundary Processing.}
In diffusion models, generated images $\hat{x_0}$ may exceed the valid pixel range. To prevent this, we apply the boundary processing function \(\varrho(\cdot)\) from \cite{chen2024content}, which clamps pixel values within $[0, 1]$, ensuring natural image integrity. 

We also introduce the projection operator $\Pi_\kappa$ to constrain adversarial perturbations $\delta$ within a $\kappa$-ball, balancing perturbation effectiveness and imperceptibility. Inspired by prior adversarial attack methods \cite{chen2024content,dong2018boosting,linnesterov,xie2019improving,xu2024highly}, we incorporate momentum for stable optimization. Momentum $g$ is updated iteratively as follows:

\begin{equation}
g_k \leftarrow \mu \cdot g_{k-1} + \frac{\nabla_{x_T} \mathcal{L} \left( \mathcal{F}_{\theta} \left( \varrho(\hat{x_0}), y \right) \right)}{ \left\| \nabla_{x_T} \mathcal{L} \left( \mathcal{F}_{\theta} \left( \varrho(\hat{x_0}), y \right) \right) \right\|_1 },
\end{equation}
where $\mu$ is the momentum coefficient, $\mathcal{L}$ is the loss function, and $\mathcal{F}_{\theta}$ represents the neural network. The adversarial perturbation $\delta_k$ is updated as:

\begin{equation}
\delta_k \leftarrow \Pi_\kappa \left( \delta_{k-1} + \eta \cdot \mathrm{sign}(g_k) \right),
\end{equation}
with $\eta$ as the step size, and $\Pi_\kappa$ ensuring the perturbation stays within the $\kappa$-ball. This method, leveraging momentum and projection, has been shown to enhance optimization in adversarial attacks while maintaining naturalistic image appearance \cite{chen2024content,dong2018boosting,linnesterov,xie2019improving,xu2024highly}.   

\section{Experiments}

\begin{table}[t]
\small
\caption{Image semantic consistency assessment}
\label{table3}
\begin{center}
\begin{tabular}{lllll}
\toprule[1.5pt]
\textbf{Attack}    & \textbf{\begin{tabular}[c]{@{}l@{}}CLIP\\ Score\end{tabular}}$\uparrow$ & \textbf{SSIM}$\uparrow$ & \textbf{PSNR}$\uparrow$ & \textbf{LPIPS}$\downarrow$ \\ \hline
SAE\cite{hosseini2018semantic}                & 0.7947                                                             & 0.9492             & 15.6864             & 0.3767       \\
ADef\cite{alaifari2018adef}               & 0.7856                                                             & 0.9112             &   14.7823           & 0.3996      \\
cAdv\cite{bhattad2019unrestricted}               & 0.8398                                                             & \textbf{0.9924}             & 19.9874             & 0.2695        \\
tAdv\cite{bhattad2019unrestricted}               & 0.8366                                                             &  0.9876            &    19.0012          & 0.2377         \\
ACE\cite{zhao2020adversarial}                & 0.8318                                                             & 0.9891             & 19.7205             &   0.1799       \\
ColorFool\cite{shamsabadi2020colorfool}          & 0.8448                                                             & 0.9358             & 9.7676             & 0.5102           \\
NCF\cite{yuan2022natural}                & 0.8848                                                             & 0.8135             & 14.3670             & 0.3194           \\
AdvST\cite{10292904} & 0.8752 &0.8144 &14.2011 &0.2908 \\
ACA\cite{chen2024content}                & 0.7442                                                             & 0.5137             & 11.0014             & 0.5971             \\ \hline
\textbf{SCA(Ours)} & \textbf{0.8896}                                               & 0.8162    & \textbf{20.5825}    & \textbf{0.1360}  \\
\bottomrule[1.5pt]
\end{tabular}
\end{center}
\end{table}

\begin{table*}[t]
\caption{Performance comparison on normally trained CNNs and ViTs. We report attack success rates ($\%$) of each method (“*” means white-box attack results).{\label{table4}}}
\centering
\resizebox{\textwidth}{!}{
\begin{tabular}{ccccccccccccc}
\toprule[1.5pt]
\multirow{3}{*}{\begin{tabular}[c]{@{}c@{}}Surrogate\\ \\ Model\end{tabular}} & \multirow{3}{*}{Attack} & \multicolumn{10}{c}{\begin{tabular}[c]{@{}c@{}}Target\\ Models\end{tabular}}                                                                                      & \multirow{3}{*}{\begin{tabular}[c]{@{}c@{}}Avg.\\ ASR(\%)\end{tabular}} \\ \cline{3-12}
                                                                              &                         & \multicolumn{6}{c}{CNNs}                                                                        & \multicolumn{4}{c}{Transformers}                                &                                                                         \\ \cline{3-12}
                                                                              &                         & MN-v2          & Inc-v3        & RN-50          & Dense-161     & RN-152        & EF-b7         & MobViT-s       & ViT-B          & Swin-B        & PVT-v2        &                                                                         \\ \hline
-                                                                             & Clean                   & 12.1           & 4.8           & 7.0            & 6.3           & 5.6           & 8.7           & 7.8            & 8.9            & 3.5           & 3.6           & 6.83                                                                    \\ \hline
\multirow{9}{*}{MobViT-s}                                                     & SAE                     & 60.2           & 21.2          & 54.6           & 42.7          & 44.9          & 30.2          & 82.5*          & 38.6           & 21.1          & 20.2          & 37.08                                                                   \\
                                                                              & ADef                    & 14.5           & 6.6           & 9.0            & 8.0           & 7.1           & 9.8           & 80.8*          & 9.7            & 5.1           & 4.6           & 8.27                                                                    \\
                                                                              & cAdv                    & 41.9           & 25.4          & 33.2           & 31.2          & 28.2          & 34.7          & 84.3*          & 32.6           & 22.7          & 22.0          & 30.21                                                                   \\
                                                                              & tAdv                    & 33.6           & 18.8          & 22.1           & 18.7          & 18.7          & 15.8          & 97.4*          & 15.3           & 11.2          & 13.7          & 18.66                                                                   \\
                                                                              & ACE                     & 30.7           & 9.7           & 20.3           & 16.3          & 14.4          & 13.8          & \textbf{99.2*} & 16.5           & 6.8           & 5.8           & 14.92                                                                   \\
                                                                              & ColorFool               & 47.1           & 12.0          & 40.0           & 28.1          & 30.7          & 19.3          & 81.7*          & 24.3           & 9.7           & 10.0          & 24.58                                                                   \\
                                                                              & NCF                     & \textbf{67.7}  & 31.2          & 60.3           & 41.8          & 52.2          & 32.2          & 74.5*          & 39.1           & 20.8          & 23.1          & 40.93                                                                   \\
                                                                               & AdvST                     & 67.2  & 44.0          &  61.1         & 46.8          & 52.4          & 33.1          & 82.3*           & 39.8           & 21.5          & 24.0          & 47.22                                                                   \\
                                                                              & ACA                     & 66.2           & 56.6          & 60.6           & 58.1          & \textbf{55.9} & 55.5          & 89.8*          & 51.4           & \textbf{52.7} & 55.1          & 56.90                                                                   \\ \cline{2-13} 
                                                                              & \textbf{SCA(Ours)}      & 65.8           & \textbf{58.9} & \textbf{61.2}  & \textbf{59.0} & 54.7          & \textbf{58.2} & 91.4*          & \textbf{52.1}  & 51.9          & \textbf{55.6} & \textbf{60.88}                                                          \\ \hline
\multirow{9}{*}{MN-v2}                                                        & SAE                     & 90.8*          & 22.5          & 53.2           & 38.0          & 41.9          & 26.9          & 44.6           & 33.6           & 16.8          & 18.3          & 32.87                                                                   \\
                                                                              & ADef                    & 56.6*          & 7.6           & 8.4            & 7.7           & 7.1           & 10.9          & 11.7           & 9.5            & 4.5           & 4.5           & 7.99                                                                    \\
                                                                              & cAdv                    & 96.6*          & 26.8          & 39.6           & 33.9          & 29.9          & 32.7          & 41.9           & 33.1           & 20.6          & 19.7          & 30.91                                                                   \\
                                                                              & tAdv                    & \textbf{99.9*} & 27.2          & 31.5           & 24.3          & 24.5          & 22.4          & 40.5           & 16.1           & 15.9          & 15.1          & 24.17                                                                   \\
                                                                              & ACE                     & 99.1*          & 9.5           & 17.9           & 12.4          & 12.6          & 11.7          & 16.3           & 12.1           & 5.4           & 5.6           & 11.50                                                                   \\
                                                                              & ColorFool               & 93.3*          & 9.5           & 25.7           & 15.3          & 15.4          & 13.4          & 15.7           & 14.2           & 5.9           & 6.4           & 13.50                                                                   \\
                                                                              & NCF                     & 93.2*          & 33.6          & \textbf{65.9}  & 43.5          & \textbf{56.3} & 33.0          & 52.6           & 35.8           & 21.2          & 20.6          & 40.28                                                                   \\
                                                                               & AdvST                     & 97.0*  & 35.4          &   57.2        & 48.5          & 50.5          & 32.4          & 52.6           & 36.8           & 19.8          & 21.7          & 45.19                                                                   \\
                                                                              & ACA                     & 93.1*          & 56.8          & 62.6           & \textbf{55.7} & 56.0          & \textbf{51.0} & 59.6           & 48.7           & \textbf{48.6} & 50.4          & 54.38                                                                   \\ \cline{2-13} 
                                                                              & \textbf{SCA(Ours)}      & 93.0*          & \textbf{56.9} & 63.3           & 53.2          & 56.0          & \textbf{51.0} & \textbf{62.3}  & \textbf{50.1}  & 47.0          & \textbf{51.0} & \textbf{58.38}                                                          \\ \hline
\multirow{9}{*}{RN-50}                                                        & SAE                     & 63.2           & 25.9          & 88.0*          & 41.9          & 46.5          & 28.8          & 45.9           & 35.3           & 20.3          & 19.6          & 36.38                                                                   \\
                                                                              & ADef                    & 15.5           & 7.7           & 55.7*          & 8.4           & 7.8           & 11.4          & 12.3           & 9.2            & 4.6           & 4.9           & 9.09                                                                    \\
                                                                              & cAdv                    & 44.2           & 25.3          & 97.2*          & 36.8          & 37.0          & 34.9          & 40.1           & 30.6           & 19.3          & 20.2          & 32.04                                                                   \\
                                                                              & tAdv                    & 43.4           & 27.0          & 99.0*          & 28.8          & 30.2          & 21.6          & 35.9           & 16.5           & 15.2          & 15.1          & 25.97                                                                   \\
                                                                              & ACE                     & 32.8           & 9.4           & \textbf{99.1*} & 16.1          & 15.2          & 12.7          & 20.5           & 13.1           & 6.1           & 5.3           & 14.58                                                                   \\
                                                                              & ColorFool               & 41.6           & 9.8           & 90.1*          & 18.6          & 21.0          & 15.4          & 20.4           & 15.4           & 5.9           & 6.8           & 17.21                                                                   \\
                                                                              & NCF                     & \textbf{71.2}  & 33.6          & 91.4*          & 48.5          & 60.5          & 32.4          & 52.6           & 36.8           & 19.8          & 21.7          & 41.90                                                                   \\
                                                                              & AdvST                     & 55.9  & 38.2          & 90.1*          & 50.5          & 57.2          & 41.3          & 52.3           & 32.8           & 20.1          & 24.7          & 46.31                                                                   \\
                                                                              & ACA                     & 69.3           & \textbf{61.6} & 88.3*          & \textbf{61.9} & 61.7          & 60.3          & \textbf{62.6}  & 52.9           & 51.9          & 53.2          & 59.49                                                                   \\ \cline{2-13} 
                                                                              & \textbf{SCA(Ours)}      & 69.5           & 60.8          & 94.5*          & 59.9          & \textbf{62.0} & \textbf{60.6} & 61.9           & \textbf{53.2}  & \textbf{52.0} & \textbf{54.1} & \textbf{62.55}                                                          \\ \hline
\multirow{9}{*}{ViT-B}                                                        & SAE                     & 54.5           & 26.9          & 49.7           & 38.4          & 41.4          & 30.4          & 46.1           & 78.4*          & 19.9          & 18.1          & 36.16                                                                   \\
                                                                              & ADef                    & 15.3           & 8.3           & 9.9            & 8.4           & 7.6           & 12.0          & 12.4           & 81.5*          & 5.3           & 5.5           & 9.41                                                                    \\
                                                                              & cAdv                    & 31.4           & 27.0          & 26.1           & 22.5          & 19.9          & 26.1          & 32.9           & 96.5*          & 18.4          & 16.9          & 24.58                                                                   \\
                                                                              & tAdv                    & 39.5           & 22.8          & 25.8           & 23.2          & 22.3          & 20.8          & 34.1           & 93.5*          & 16.3          & 15.3          & 24.46                                                                   \\
                                                                              & ACE                     & 30.9           & 11.4          & 22.0           & 15.5          & 15.2          & 13.0          & 17.0           & 98.6*          & 6.5           & 6.3           & 15.31                                                                   \\
                                                                              & ColorFool               & 45.3           & 13.9          & 35.7           & 24.3          & 28.8          & 19.8          & 27.0           & 83.1*          & 8.9           & 9.3           & 23.67                                                                   \\
                                                                              & NCF                     & 55.9           & 25.3          & 50.6           & 34.8          & 42.3          & 29.9          & 40.6           & 81.0*          & 20.0          & 19.1          & 35.39                                                                   \\
                                                                               & AdvST                     & 53.9  & 28.6          & 52.1          & 38.5          & 40.5          & 32.4          & 42.2           & 85.1*          & 19.8          & 21.7          & 41.48                                                                   \\
                                                                              & ACA                     & \textbf{64.6}  & 58.8          & \textbf{60.2}  & \textbf{58.1} & \textbf{58.1} & \textbf{57.1} & \textbf{60.8}  & 87.7*          & 55.5          & \textbf{54.9} & 58.68                                                                   \\ \cline{2-13} 
                                                                              & \textbf{SCA(Ours)}      & 64.1           & \textbf{59.0} & 60.0           & 57.9          & 58.0          & 57.0          & 60.6           & \textbf{89.0*} & \textbf{58.1} & 54.0          & \textbf{61.77}                                                          \\ 
\bottomrule[1.5pt]
\end{tabular}
}
\end{table*}

\subsection{Experimental Setup}
\textbf{Datasets.} We carry out our experiments using the ImageNet-compatible dataset \cite{kurakin2018adversarial}, which includes 1,000 images from the ImageNet validation set \cite{deng2012imagenet}. This dataset has been frequently employed in related work \cite{dong2019evading,gao2020patch,xie2019improving,yuan2022natural,chen2024content}.

\textbf{Semantic Consistency Evaluation.} We use CLIP Score \cite{radford2021learning}, SSIM \cite{wang2004image}, PSNR \cite{huynh2008scope}, and LPIPS \cite{zhang2018unreasonable} to measure the semantic consistency between the generated unrestricted adversarial examples and the clean image from multiple aspects.

\textbf{Attack Evaluation.} We compare SCA with other unrestricted adversarial attacks, including SAE \cite{hosseini2018semantic}, ADef \cite{alaifari2018adef}, cAdv \cite{bhattad2019unrestricted}, tAdv \cite{bhattad2019unrestricted}, ACE \cite{zhao2020adversarial}, ColorFool \cite{shamsabadi2020colorfool}, NCF \cite{yuan2022natural}, AdvST \cite{10292904}, and ACA \cite{chen2024content}. The default parameter settings for these attacks are used. The evaluation metric employed is attack success rate (ASR), which measures the proportion of images misclassified by the target models.

\textbf{Target Models.} To assess the adversarial robustness of various architectures, we target both convolutional neural networks (CNNs) and vision transformers (ViTs). The CNN models include MobileNet-V2 (MN-v2), Inception-v3 (Inc-v3), ResNet-50 (RN-50), ResNet-152 (RN-152), DenseNet-161 (Dense-161), and EfficientNet-b7 (EFb7). For ViTs, we evaluate MobileViT (MobViT-s), Vision Transformer (ViT-B), Swin Transformer (Swin-B), and Pyramid Vision Transformer (PVT-v2).

\textbf{Implementation Details.} All experiments are performed using PyTorch on an NVIDIA Tesla A100. The DDPM steps are set to $T = 10\to20$, with attack iterations $N_a = 10$, $\eta = 0.04$, $\kappa = 0.1$, and $\mu=1$. The version of Stable Diffusion \cite{rombach2022high} used is v1.5, and image captions are generated automatically via LLaVA-NeXT.

\subsection{Semantic Consistency Comparison}
To comprehensively evaluate the semantic consistency preservation ability of current unrestricted adversarial
attack methods from multiple perspectives, we introduce four
indicators, namely CLIP Score, SSIM, PSNR, and MSE. The specific results are shown in Table~\ref{table3}. PSNR (Peak Signal to
Noise Ratio) is a metric for evaluating image quality. The unit
of PSNR is dB. The larger the value, the less image distortion.
However, many experimental results show that the PSNR score
cannot be completely consistent with the visual quality seen
by the human eye. A higher PSNR image may look worse than a lower PSNR image. This is because the
sensitivity of the human eye to errors is not absolute, and its
perception results will vary due to many factors. To
address the limitations of PSNR, we introduced LPIPS and
SSIM. The range of SSIM values is [0,1], and the larger
the value, the more similar the images are. If two images
are the same, the SSIM value is 1. LPIPS (Learned
Perceptual Image Patch Similarity), also known as perceptual
loss, is used to measure the difference between two images.
This metric learns the reverse mapping from the generated
image to the ground truth, forces the generator to learn the
reverse mapping of reconstructing the real image from the fake
image, and prioritizes the perceptual similarity between them.
LPIPS is more in line with human perception than traditional
methods. The lower the LPIPS value, the more similar the two
images are, and vice versa, the greater the difference.

In addition, we also introduced CLIP Score as a more
advanced semantic evaluation metric. As we all know, CLIP
has been trained on more than 400 million image-text pairs and
has strong understanding capabilities. So we measure semantic
consistency by calculating the cosine similarity between the
image embeddings processed by CLIP’s visual encoder.

In summary, the experimental results show that the adversarial examples we generate outperform existing methods in most
indicators, especially in LPIPS and CLIP Score. This further
proves the superiority of our SCA in maintaining semantic
consistency.

\begin{figure*}[t]
\begin{center}
\subcaptionbox{Comparison on an Image of a Cup}{
\subcaptionbox*{clean}{\includegraphics[width = 0.105\textwidth]{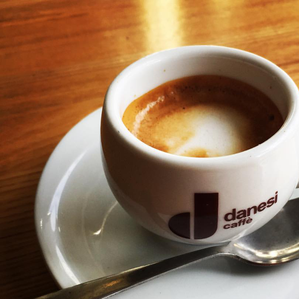}}
	\subcaptionbox*{cAdv}{\includegraphics[width = 0.105\textwidth]{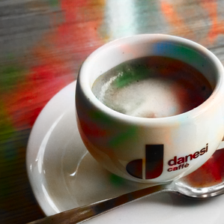}}

	\subcaptionbox*{SAE}{\includegraphics[width = 0.105\textwidth]{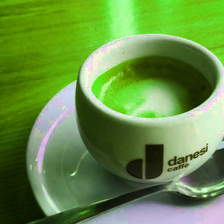}} 
 \subcaptionbox*{ADef}{\includegraphics[width = 0.105\textwidth]{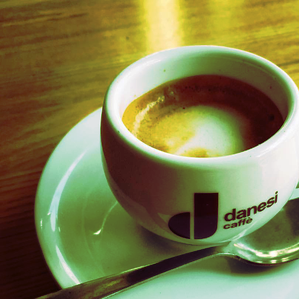}} 
 \subcaptionbox*{ColorFool}{\includegraphics[width = 0.105\textwidth]{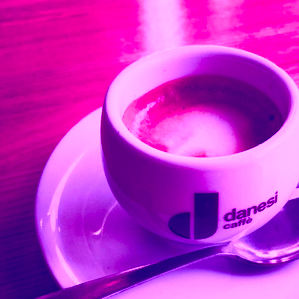}}

	\subcaptionbox*{NCF}{\includegraphics[width = 0.105\textwidth]{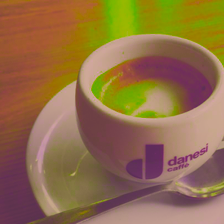}}

  	\subcaptionbox*{AdvST}{\includegraphics[width = 0.105\textwidth]{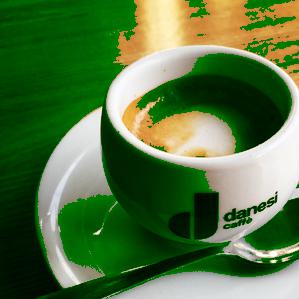}}

	\subcaptionbox*{ACA}{\includegraphics[width = 0.105\textwidth]{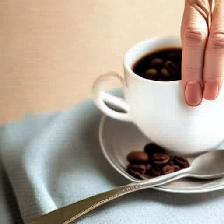}} 
 \subcaptionbox*{\textbf{Ours}}{\includegraphics[width = 0.105\textwidth]{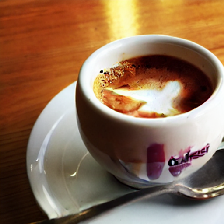}}}
 \subcaptionbox{Comparison on an Outdoor Environment Photo}{
 \subcaptionbox*{clean}{\includegraphics[width = 0.105\textwidth]{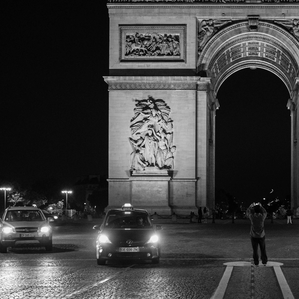}}
	\hfill
	\subcaptionbox*{cAdv}{\includegraphics[width = 0.105\textwidth]{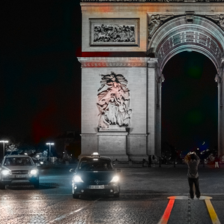}}
	\hfill
	\subcaptionbox*{SAE}{\includegraphics[width = 0.105\textwidth]{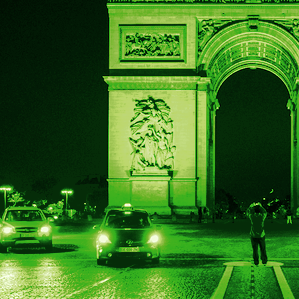}} 
 \subcaptionbox*{ADef}{\includegraphics[width = 0.105\textwidth]{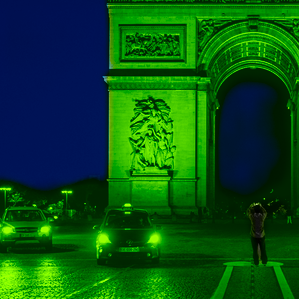}} 
 \subcaptionbox*{ColorFool}{\includegraphics[width = 0.105\textwidth]{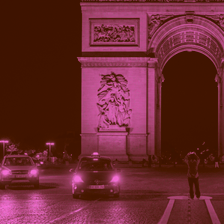}}
	\hfill
	\subcaptionbox*{NCF}{\includegraphics[width = 0.105\textwidth]{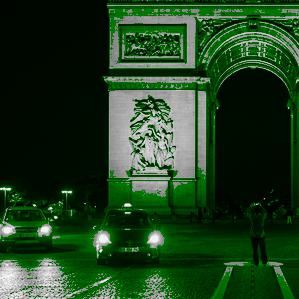}}
	\hfill
 	\subcaptionbox*{AdvST}{\includegraphics[width = 0.105\textwidth]{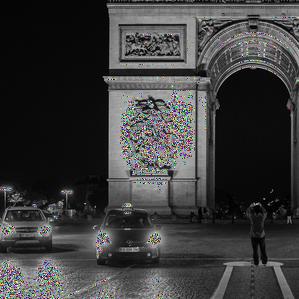}}
	\hfill
	\subcaptionbox*{ACA}{\includegraphics[width = 0.105\textwidth]{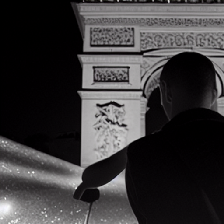}} 
 \subcaptionbox*{\textbf{Ours}}{\includegraphics[width = 0.105\textwidth]{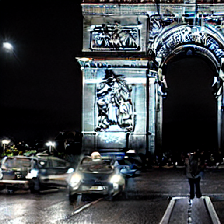}}}
 \subcaptionbox{More results of our SCA}{
 \subcaptionbox*{clean image}{\includegraphics[width = 0.115\textwidth]{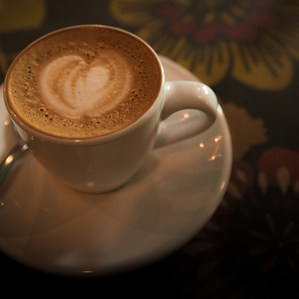}}
	\hfill
	\subcaptionbox*{Adv}{\includegraphics[width = 0.115\textwidth]{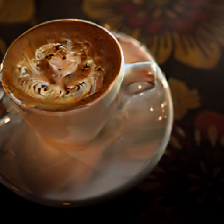}}
	\hfill
	\subcaptionbox*{clean image}{\includegraphics[width = 0.115\textwidth]{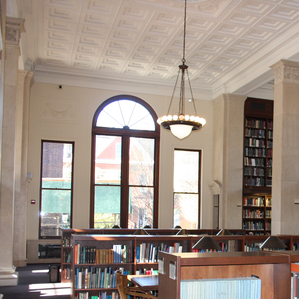}} 
 \subcaptionbox*{Adv}{\includegraphics[width = 0.115\textwidth]{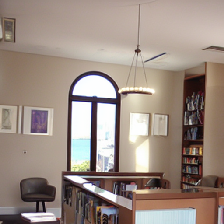}} 
 \subcaptionbox*{clean image}{\includegraphics[width = 0.115\textwidth]{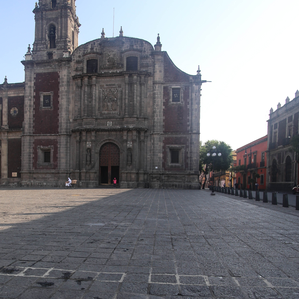}}
	\hfill
	\subcaptionbox*{Adv}{\includegraphics[width = 0.115\textwidth]{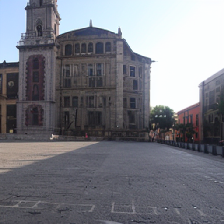}}
	\hfill
 	\subcaptionbox*{clean image}{\includegraphics[width = 0.115\textwidth]{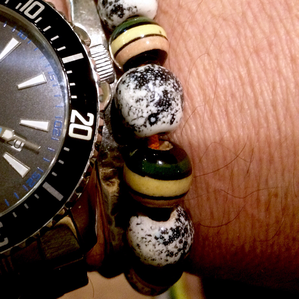}}
	\hfill
	\subcaptionbox*{Adv}{\includegraphics[width = 0.115\textwidth]{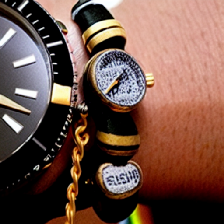}} 
 }
\end{center}
   \caption{Compared with other attacks, SCA generates the most natural adversarial examples and maintains a high degree of semantic consistency with the clean image. }
\label{fig:3}
\end{figure*}

\subsection{Comparison of Attack Success Rates}
We systematically evaluate the adversarial robustness of convolutional neural networks (CNNs) and vision transformers (ViTs) using several prominent adversarial methods: SAE \cite{hosseini2018semantic}, ADef \cite{alaifari2018adef}, cAdv and tAdv \cite{bhattad2019unrestricted}, ACE \cite{zhao2020adversarial}, ColorFool \cite{shamsabadi2020colorfool}, NCF \cite{yuan2022natural}, AdvST \cite{10292904}, ACA \cite{chen2024content}, and our proposed SCA. These methods were applied to MobileNetV2 (MN-v2), ResNet-50 (RN-50), MobViT-s, and ViT-B, with performance measured using the Average Attack Success Rate (\%), which captures the transferability of adversarial examples across non-substitute models.

Table~\ref{table4} provides a detailed comparison of these methods in transferring adversarial examples between CNNs and ViTs. Our SCA consistently achieves high attack success rates, often surpassing or equaling ACA, a leading method in the field. Notably, SCA demonstrates slight but meaningful improvements over ACA when using MN-v2 as a surrogate model, particularly on Inception-v3 (Inc-v3), RN-50, and MobViT-s, emphasizing SCA’s superior adversarial transferability.

Our results confirm that SCA achieves state-of-the-art performance by enhancing the imperceptibility and stealthiness of adversarial examples through semantic consistency. Crucially, this semantic alignment does not compromise attack efficacy; instead, it enhances transferability without sacrificing performance.

Both SCA and ACA show strong transferability across architectures, particularly in the challenging case of transferring adversarial examples from CNNs to ViTs. This is due to two factors: (i) both methods perform a low-dimensional manifold search within the natural image space, inherently improving cross-architecture transferability, and (ii) both leverage diffusion models with self-attention mechanisms that promote architectural similarities. With an average ASR exceeding 60\%, SCA not only advances the state of the art but also provides an effective approach for generating transferable adversarial examples across diverse model architectures, demonstrating its robustness and efficiency.

\begin{table}[t]
\small
\caption{Image quality assessment{\label{table2}}}
\begin{center}
\begin{tabular}{llllll}
\toprule[1.5pt]
Attack    & \begin{tabular}[c]{@{}l@{}}NIMA\\ -AVA\end{tabular} & \begin{tabular}[c]{@{}l@{}}Hyper \\-IQA\end{tabular} & \begin{tabular}[c]{@{}l@{}}MUSIQ\\ -AVA\end{tabular} & \begin{tabular}[c]{@{}l@{}}MUSIQ\\ -KonIQ\end{tabular} & TReS  \\ \hline
Clean     & 5.15                                                & 0.667    & 4.07                                                 & 52.66                                                  & 82.01 \\ \hline
SAE       & 5.05                                                & 0.597    & 3.79                                                 & 47.24                                                  & 71.88 \\
ADef      & 4.89                                                & 0.608    & 3.89                                                 & 47.39                                                  & 72.10 \\
cAdv      & 4.97                                                & 0.623    & 3.87                                                 & 48.32                                                  & 75.12 \\
tAdv      & 4.83                                                & 0.525    & 3.78                                                 & 44.71                                                  & 67.07 \\
ACE       & 5.12                                                & 0.648    & 3.96                                                 & 50.49                                                  & 77.25 \\
ColorFool & 5.24                                                & 0.662    & 4.05                                                 & 52.27                                                  & 78.54 \\
NCF       & 4.96                                                & 0.634    & 3.87                                                 & 50.33                                                  & 74.10 \\
AdvST       & 5.12                                                & 0.679    & 3.99                                                 & 56.33                                                  & 79.21 \\
ACA       & \textbf{5.54}                                                & \textbf{0.691}    & 4.37                                                 & 56.08                                                  & \textbf{85.11} \\ \hline
SCA(Ours) & 5.32                                                   & \textbf{0.691}        & \textbf{4.47}                                                    & \textbf{58.62}                                                      & 82.05  \\
\bottomrule[1.5pt]
\end{tabular}
\end{center}
\end{table}

\subsection{Image Quality Comparison.}

In this section, we conduct a detailed quantitative assessment of the image quality for adversarial examples, following the approach taken in prior works such as \cite{shamsabadi2020colorfool}, \cite{yuan2022natural}, and \cite{chen2024content}. Specifically, we employ non-reference perceptual image quality metrics to evaluate the visual realism and aesthetic appeal of the adversarial images generated by different methods. The metrics we use include NIMA \cite{talebi2018nima}, HyperIQA \cite{su2020blindly}, MUSIQ \cite{ke2021musiq}, and TReS \cite{golestaneh2022no}, which are widely recognized for their ability to assess image quality in the absence of a ground truth reference.

To provide a robust evaluation, we follow the PyIQA framework and apply models trained on large-scale aesthetic datasets. NIMA-AVA and MUSIQ-AVA are trained on the AVA \cite{murray2012ava} dataset, while MUSIQ-KonIQ is trained on the KonIQ-10K \cite{hosu2020koniq} dataset, which ensures a comprehensive evaluation of the generated adversarial examples. As shown in Table~\ref{table2}, our method consistently achieves image quality scores that are comparable to ACA, outperforming other methods such as ColorFool and NCF. This superior performance is likely attributable to the fact that both SCA and ACA leverage Stable Diffusion, a state-of-the-art generative model known for its ability to produce highly realistic and visually appealing images.

The use of Stable Diffusion plays a critical role in enhancing the naturalness and aesthetic quality of our adversarial examples. The inherent generative power of this model allows us to craft perturbations that blend seamlessly into the clean image, preserving its photorealism while embedding subtle adversarial signals. Furthermore, we map the clean image to a latent space using an inversion method, which effectively places the image within a low-dimensional manifold representing the distribution of natural images. By perturbing the image within this manifold, our method ensures that the adversarial examples remain visually coherent and natural, aligning closely with the underlying distribution of real-world images.

Additionally, it is worth noting that the no-reference image quality metrics we employ are often trained on aesthetic datasets, many of which contain post-processed images, including those edited with tools such as Photoshop. These post-processed images are more aligned with human aesthetic preferences, which is an important consideration for assessing adversarial examples. As \cite{chen2024content} points out, the similarity of our method to post-processing and image-editing techniques helps explain why our adversarial examples receive higher quality scores from these metrics. Our approach effectively functions as a form of image editing, subtly modifying the image in ways that retain or even enhance its aesthetic appeal, leading to better evaluation results.

\subsection{Qualitative Comparison}

In this section, we present a qualitative analysis of various current unrestricted attack methods, illustrated in Figure~\ref{fig:3}. Our findings reveal that the SCA generates adversarial examples that exhibit a greater degree of naturalness compared to methods that rely heavily on color and texture transformations. Notably, techniques such as SAE and ColorFool often produce adversarial examples characterized by unnatural color distortions, making them easily detectable to human observers. This perceptibility undermines their effectiveness as adversarial attacks.

In contrast, the state-of-the-art methods ACA and AdvST also achieve a commendable level of naturalness in their generated adversarial examples. However, when it comes to preserving semantic consistency with clean images, our method stands out. For instance, as depicted in Figure~\ref{fig:3}(a), the adversarial image produced by our approach only alters the pattern of the coffee cup while retaining the overall environmental context of the scene. This subtle manipulation contrasts sharply with ACA, which, while maintaining the basic concept of the coffee cup, leads to significant alterations in both the cup's shape and the surrounding environment. Such transformations diminish the connection to the original clean image, resulting in a lack of semantic fidelity.

\begin{table*}[t]
\small
\caption{Attack speed of unrestricted attacks. We choose MN-v2 as the surrogate model and evaluate the inference time on an NVIDIA Tesla A100.{\label{table1}}}
\begin{center}
\begin{tabular}{c c c c c c c c c c c c}
\toprule[1.5pt]
Attack & SAE  &ADef& cAdv & tAdv & ACE & ColorFool & NCF &AdvST & ACA &  \textbf{Ours(SCA)}\\
\hline
Time(Sec) & 8.80 & 0.41&18.67 & 4.88 & 6.64 & 12.18 & 10.45 &19.25 & 125.33 & 9.85 \\
\bottomrule[1.5pt] 
\end{tabular}
\end{center}
\end{table*}

\begin{table*}[t]
\caption{Ablation study of random gradient-free method.{\label{last}}}
\centering
\begin{tabular}{ccccccc}
\toprule[1.5pt]
\multicolumn{1}{c}{\multirow{2}{*}{Methods}} & \multicolumn{5}{c}{Models}                                                     & \multicolumn{1}{c}{\multirow{2}{*}{\begin{tabular}[c]{@{}c@{}}Avg.\\ ASR\end{tabular}}} \\ \cline{2-6}
\multicolumn{1}{c}{}                         & Dense161      & RN-152        & EF-b7         & ViT-B          & PVT-v2        & \multicolumn{1}{c}{}                                                                    \\ \hline
\multicolumn{1}{c}{\begin{tabular}[c]{@{}c@{}}Clean\\Image\end{tabular}}              & 6.3           & 5.6           & 7.7           & 8.9*            & 3.6           & 6.62                                                                                    \\ \hline
\multicolumn{1}{c}{NE}            & 7.0           & 6.2           & 9.1           & 12.3*           & 5.1           & 7.94                                                                                    \\
\multicolumn{1}{c}{SG}                       & 24.3          & 22.3          & 27.5          & 59.6*           & 19.7          & 30.68                                                                                   \\
\multicolumn{1}{c}{\textbf{RGF}}             & \textbf{57.9} & \textbf{58.0} & \textbf{57.0} & \textbf{89.0*} & \textbf{54.0} & \textbf{63.18}                                                                          \\ 
\bottomrule[1.5pt]
\end{tabular}
\end{table*}

\subsection{Time Analysis}
In this section, we present a comparison of the attack speeds for various unrestricted attacks. Using MN-v2 as the surrogate model, we measure the inference time on an NVIDIA Tesla A100. Table~\ref{table1} shows the average time (in seconds) required to generate an adversarial example per image. Compared with the current state-of-the-art methods, we have increased the speed by about 12 times while maintaining the diversity of generated content and the success rate of attacks. This is mainly due to our novel inversion method, and the use of DPM Solver++ greatly accelerates the Diffusion sampling process. Experiments have shown that we can reduce the number of sampling steps to 10 to 20 steps without affecting the quality of the generated images. From Table~\ref{table1}, it can be seen that some methods take a very short time, but experiments have shown that their effects are generally poor. Overall, our method is the most efficient.

\subsection{Ablation Study}
In the process of computing the gradients for optimization based on the gradient optimization process using Equation~(\ref{eq:10}) and Equation~(\ref{eq:11}), we apply the random gradient-free (RGF) method. To evaluate its effectiveness, we conducted ablation tests under three conditions: without gradient estimation (NE), using the Skip Gradient method (SG), and using RGF. Table~\ref{last} shows the results. Our SCA performs poorly without gradient estimation and with the Skip Gradient method. The ASR for ``No estimation'' is similar to the clean image, showing that we can't ignore the diffusion model's sampling and directly optimize with the classifier's loss in the latent space. The Skip Gradient method also underperforms, likely due to the need for more sample steps, as we use complex optimizations to speed up sampling. Additionally, our new inversion method differs from prior approximations, validating the importance of the RGF method in our framework.

\begin{figure}
\begin{center}
\subfloat[Null]{\includegraphics[width = 0.3\textwidth]{Figures/image5.1.png}}
	\hfill
	\subfloat[BLIP-2]{\includegraphics[width = 0.3\textwidth]{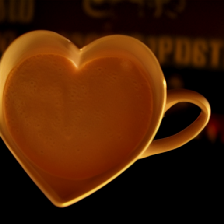}}
	\hfill
	\subfloat[LLaVA-NeXT]{\includegraphics[width = 0.3\textwidth]{Figures/image5.2.png}} 
\caption{Ablation study of the degree of semantic restriction. The caption of (b) is ``a coffee cup with a heart-shaped design in the foam''. The caption of (c) is ``At the center of the frame is a white cup filled with a dark brown liquid, possibly coffee. The cup is placed on a white saucer, which is accompanied by a silver spoon. All these items rest on a tablecloth featuring a colorful floral pattern. On the cup, there's a heart-shaped latte art design, suggesting a level of care and attention given to the preparation of the drink.''}
\label{figlast}
\end{center}
\end{figure}

We also tested the impact of using MLLM models for semantic consistency. In this experiment, we replaced LLaVA-NexT with BLIP-2 for generating captions during perturbation and also tried unconditional generation without prompts. As shown in Figure~\ref{figlast}, BLIP-2 produces very brief descriptions, leading to a significant semantic drift (Figure~\ref{figlast}(b)). In contrast, LLaVA-NexT generates detailed captions, maintaining better semantic alignment with the clean image (Figure~\ref{figlast}(c)). Additional experiments confirm that richer captions improve semantic preservation. In the case of unconditional generation (Figure~\ref{figlast}(a)), the output remains nearly identical to the original, failing to produce adversarial perturbations, likely due to overly strong semantic constraints. This highlights the importance of balancing semantic restriction, an area that warrants further research.

\section{Conclusion}
In this paper, We propose a novel attack framework called Semantic-Consistent Unrestricted Adversarial Attack (SCA) via Semantic Fixation Inversion and Semantically Guided Perturbation. The core idea of SCA is to enhance semantic control throughout the entire generation process of unrestricted adversarial examples. We initially utilized an effective inversion method and a powerful MLLM to extract rich semantic priors from a clean image. Subsequently, we optimize the adversarial objective within the latent space under the guidance of these priors. Experiments demonstrate that the adversarial examples exhibit a high degree of semantic consistency compared to existing methods. Furthermore, our method is highly efficient. Consequently, we introduce Semantic-Consistent Adversarial Examples (SCAE). Our work can shed light on further understanding the vulnerabilities of DNNs as well as novel defense approaches.

\bibliographystyle{plain}
\bibliography{main}

\end{document}